%% file: bare_jrnl.tex
\documentclass[journal]{IEEEtran}

\usepackage[pdftex]{graphicx}
\graphicspath{{images/}}
\DeclareGraphicsExtensions{.eps,.png}
\usepackage{comment}
\usepackage{cite}
\usepackage{url}
\usepackage{amsmath}
\usepackage{amssymb}
\usepackage{algorithm}
\usepackage[noend]{algpseudocode}
\algnewcommand\algorithmicforeach{\textbf{for each}}
\algdef{S}[FOR]{ForEach}[1]{\algorithmicforeach\ #1\ \algorithmicdo}
\usepackage{tabularx,textcomp}
\usepackage{multirow}
\usepackage{makecell}
\usepackage{subcaption}
\newtheorem{definition}{Definition}
\DeclareMathOperator{\sign}{sign}

\newcommand{\added}[1]{#1}

\begin{document}
\title{SeReNe: Sensitivity based  Regularization  of  Neurons for Structured Sparsity in  Neural Networks}


\author{Enzo~Tartaglione~\IEEEmembership{Member,~IEEE,}
        Andrea~Bragagnolo,
        Francesco~Odierna,\\
        Attilio Fiandrotti,~\IEEEmembership{Senior Member,~IEEE,}
        and~Marco~Grangetto,~\IEEEmembership{Senior Member,~IEEE}
\thanks{E. Tartaglione, A. Bragagnolo, F. Odierna, A. Fiandrotti and M. Grangetto are with the Computer Science Department, University of Turin, Torino, ITALY, e-mail: first.last@unito.it}
\thanks{A. Fiandrotti is also with LTCI, T\'el\'ecom Paris, Institut Polytechnique de Paris, FRANCE, e-mail:attilio.fiandrotti@telecom-paris.fr}
}

\maketitle

\begin{abstract}
Deep neural networks include millions of learnable parameters, making their deployment over resource-constrained devices problematic.
SeReNe (Sensitivity-based  Regularization  of  Neurons) is a method for learning sparse topologies with a structure, exploiting \textit{neural sensitivity} as a regularizer.
We define the sensitivity of a neuron as the variation of the network output with respect to the variation of the activity of the neuron.
The lower the sensitivity of a neuron, the less the network output is perturbed if the neuron output changes.
By including the neuron sensitivity in the cost function as a regularization term, we are able to prune neurons with low sensitivity.
As entire neurons are pruned rather then single parameters, practical network footprint reduction becomes possible. 
Our experimental results on multiple network architectures and datasets yield competitive compression ratios with respect to state-of-the-art references.
\end{abstract}


\IEEEpeerreviewmaketitle

\input{1_introduction}

\input{2_related}
\input{3_method}
\input{4_training_new}
\input{5_results}
\input{6_conclusion}

\appendices
\input{A1_upperboundeff}
\input{tables/lenet_300_sigmoid_tab}
\input{A2_Jderivation}
\input{A3_17deriv}
\input{A4_sigmoidact}



%




\ifCLASSOPTIONcaptionsoff
  \newpage
\fi



\bibliographystyle{IEEEtran}
\bibliography{bare_jrnl}
%


%








\end{document}

%% file: 1_introduction.tex

\section{Introduction}
\label{sec:intro}
\IEEEPARstart{D}{eep} Neural Networks (DNNs) can solve extremely challenging tasks thanks to complex stacks of (convolutional) layers with thousands of neurons~\cite{krizhevsky2012imagenet, simonyan:vgg, he2016deep}. 
Let us define here the \textit{complexity} of a neural network as the number of its learnable parameters: architectures such as AlexNet and VGG have a complexity in the order of 60 and 130 million parameters respectively.
Similar architectures are challenging to deploy in scenarios where resources such as the memory or storage are limited.
For example, the 8-layers AlexNet~\cite{krizhevsky2012imagenet} memory footprint exceeds 240MB of memory, whereas the 19-layers VGG-Net~\cite{simonyan:vgg} footprint exceeds 500 MB. 
The need for compact DNNs is witnessed also by the fact that the Moving Pictures Experts Group (MPEG) of ISO has recently broadened its scope beyond multimedia contents issuing an exploratory call for proposal to compress neural networks~\cite{mpegNetCompr}.
Multiple (complementary) approaches are possible to cope with neural networks memory requirements, inference time and energy consumption:

\begin{itemize}
    \item Re-designing the network topology. Moving from one architecture to another, possibly forcing a precise neuronal connectivity, or weight sharing, can reduce the number of parameters, or the complexity of the network~\cite{he2016deep, lu2019srgc}.
    \item \added{Quantization. Representing the parameters (and activation functions) as fixed-point digits reduces the memory footprint and speeds up computations~\cite{wu2016quantized}.}
    \item Pruning. Deep architectures need to be over-parametrized~\cite{sagun2017empirical} to be trained effectively, but redundant parameters can be pruned at inference time \cite{lecun1990optimal,
    kingma:var_dropout, tartaglione:sensitivity, han:deep_compression, frankle2018lottery}.
    The present work falls in this latter category.
\end{itemize}

Pruning techniques aim at learning sparse topologies by selectively dropping synapses between neurons (or neurons altogheter when all incoming synapses are dropped).
For example, \cite{tartaglione:sensitivity} and \cite{han:deep_compression} apply a regularization function promoting low magnitude weights followed by zero thresholding or quantization.
Such approaches slash the number of non-zero parameters, allowing to represent the parameters of a layer as a sparse tensor \cite{naumov2010cusparse}.
Such methods aim however at pruning parameters independently, so the learned topologies lacks a structure despite sparse.
Storing and accessing in memory a randomly sparse matrix entails significant challenges, so it is unclear to which extent such methods could be practically exploited.

This work proposes SeReNe, a method for learning sparse network topologies with a structure, i.e. with fewer neurons altogether.
In a nutshell, our method drives \textit{all} the parameters of a neuron towards zero, allowing to prune entire neurons from the network.
\\
First, we introduce the notion \textit{sensitivity of a neuron} as the variation of the network output with respect to the neuron activity. The latter is measured as the post-synaptic potential of the neuron, i.e. the input to the neuron's activation function. 
The underlying intuition is that neurons with low sensitivity yield little variation in the network output and thus negligible performance loss if their output changes locally.
We also provide computationally efficient bounds to approximate the sensitivity.
\\
Second, we design a regularizer that shrinks 
\textit{all} parameters of low sensitivity neurons, paving the way to their removal.
Indeed, when the sensitivity of a neuron approaches zero, the neuron no longer emits signals and is ready to be pruned.
\\
Third, we propose an iterative two-steps procedure to prune parameters belonging to low sensitivity neurons. Through a cross-validation strategy, we ensure controlled (or even no) performance loss with respect to the original architecture.
\\
Our method allows to learn network topologies which are not only sparse, i.e. with few non-zero parameters, but with fewer neurons (fewer filters for convolutional layers).
As a side benefit, smaller and denser architectures may also speedup network execution thanks to a better use of cache locality and memory access pattern.
\\
We experimentally show that SeReNe outperforms state-of-the-art references over multiple learning tasks and network architectures.
We observe the benefit of structured sparsity when storing the neural network topology and parameters using the \textit{Open Neural Network eXchange} format~\cite{bai2019}, with a reduction of the memory footprint.

The rest of the paper is structured as follows. In Sec.~\ref{sec:related} we review the relevant literature in neural network pruning. In Sec.~\ref{sec:NS} we provide the definition of sensitivity and practical bounds for its computation; then, we present a parameter update rule to ``drive'' the parameters of low-sensitivity neurons towards zero. Follows, in Sec.~\ref{sec:training}, a practical procedure to prune a network with our scheme. Then, in Sec.~\ref{sec::results} all the empirical results are shown and finally, in Sec.~\ref{sec:conclusion}, the conclusions are drawn.

%% file: 2_related.tex
\section{Related work}
\label{sec:related}


Approaches towards compact neural networks representations can be categorized in three major groups: altering the network structure, quantizing the parameters and pruning weights. In this section, we review works based on a pruning approach that are most relevant to our work.\\
In their seminal paper~\cite{lecun1990optimal}, LeCun~\textit{et~al.} proposed to remove unimportant weights from a network, measuring the importance of each single weight as the increment on the train error when the weight is set to zero. Unfortunately, the complexity of such method would becomes computationally unbearable in the case of deep topologies with millions of parameters.
Due to the scale and the resources required to train and deploy modern deep neural networks, sparse architectures and compression techniques have gained much interest in the deep learning community. Several successful approaches to this problem have been proposed \cite{lebedev:brain_damage, liu:sparse_cnn, zhu:pruning, wen:struct_spars}.
\added{While a more in depth analysis on the topic has been published by Gale~\textit{et~al}.~\cite{gale:state}, in the rest of this section we provide a summary of the main techniques used to prune deep architectures.}\\
\textbf{Evolutionary algorithms.} Multi-objective sparse feature learning has been proposed by Gong~\emph{et~al.}~\cite{gong2015multiobjective}: with their evolutionary algorithm, they were able to find a good compromise between sparsity and learning error, at the cost, however, of high computational cost. \added{Similar drawbacks can be found in the work by Lin~\emph{et~al.}, where convolutional layers are pruned using the artificial bee colony optimization algorithm (dubbed as ABCPruner)~\cite{lin2020channel}}.\\
\added{\textbf{Dropout.} Dropout aims at preventing a network from over-fitting by randomly dropping some neurons at learning time~\cite{srivastava:dropout}. Despite dropout tackles a different problem, it has inspired some techniques aiming at sparsifying deep architectures.} Kingma~\textit{et al.}~\cite{kingma:var_dropout} have shown that dropout can be seen as a special case of Bayesian regularization. Furthermore, they derive a variational method that allows to use dropout rates adaptively to the data. Molchanov~\textit{et~al.}~\cite{molchanov:var_dropout} exploited such variational dropout to sparsify both fully-connected and convolutional layers. In particular, the parameters having high dropout rate are always ignored and they can be removed from the network. Even if this technique obtains good performance, it is quite complex and it is reported to behave inconsistently when applied to deep architectures~\cite{gale:state}. Furthermore, this technique relies on the belief that the Bernoulli probability distribution (to be used with the dropout) is a good variational approximation for the posterior. \added{Another dropout-based approach is \emph{Targeted Dropout}~\cite{gomez2019targeted}: here, fine-tuning the ANN model is self-reinforcing its sparsity by stochastically dropping connections. They also target structured sparsity without, however, reaching state-of-the-art performance.}\\
\textbf{Knowledge distillation.} \added{Recently, knowledge distillation~\cite{hinton2015distilling} received significant attention. 
The goal in this case is to train a single network to have the same behavior (in terms of outputs under certain inputs) as an ensemble of models, reducing the overall computational complexity.
Distillation finds application in reducing the prediction of multiple networks into a single one, but can not be applied to minimize the number of neurons for a single network. \added{A recent work is \textit{Few Samples Knowledge Distillation} (FSKD)~\cite{li2020few}, where a small student network is trained from a larger teacher.} In general, in distillation-based techniques, the architecture to be trained is a-priori known, and kept static through all the learning process: in this work, we aim at providing an algorithm which automatically shrinks the deep model's size with minimal overhead introduced.}\\ 
\textbf{Few-shot pruning.} Another approach relies on defining the importance of each connection and later remove parameters deemed unnecessary. 
\added{A recent work by Frankle~and~Carbin~\cite{frankle2018lottery} proposed the \emph{lottery ticket hypothesis}, which is having a large impact on the research community. They claim that from an ANN, early in the training, it is possible to extract a sparse sub-network, using a one-shot or iterative fashion: such sparse network, when re-trained, can match the accuracy of the original model. This technique has multiple requirements, like having the history of the training process in order to detect the ``lottery winning parameters'', and it is not able to self-tune an automatic thresholding mechanism. Lots of efforts are devoted towards making pruning mechanisms more efficient: for example, Wang~et~al. show that some sparsity is achievable pruning weights at the very beginning of the training process~\cite{wang2020pruning}, or Lee~et~al., with their ``SNIP'', are able to prune weights in a one-shot fashion~\cite{Lee2019SNIPSN}. However, these approaches achieve limited sparsity: iterative pruning-based strategy, when compared to one-shot or few-shot approaches, are able to achieve a higher sparsity~\cite{tartaglione2020pruning}.}\\
\textbf{Regularization-based pruning.} \added{Finally, regularization-based approaches rely on a  regularization term (designed to enhance sparsity) to be minimized besides the loss function at training time. Louizos~\textit{et~al.} propose an $\ell_0$ regularization to prune the network parameters during training~\cite{louizos:sparse_l0}. Such a technique penalizes non-zero value of a parameter vector, promoting sparse solutions. As a drawback, it requires solving a complex optimization problem, besides the loss minimization strategy and other regularization terms.} Han~\textit{et~al.} propose a multi-step process in which the least relevant parameters are defined, minimizing a target loss function~\cite{han:deep_compression}. In particular, it relies on a thresholding heuristics, where all the less important connections are pruned. In~\cite{tartaglione:sensitivity}, a similar approach was followed, introducing a novel regularization term that measures the ``sensitivity'' of the output wrt. the variation of the parameters. While this technique achieves top-notch sparsity even in deep convolutional architectures, such sparsity is not structured, i.e. the resulting topology includes large numbers of neurons with at least one non-zero parameter. Such unstructured sparsity bloats the practically attainable network footprint and leads to irregular memory accesses, jeopardizing execution speedups. In this work we aim at overcoming the above limitations proposing a regularization method that produces a \emph{structured} sparsification, focusing on removing entire neurons instead of single parameters.
We also leverage our recent research showing that post-synaptic potential regularization is able to boost generalization over other regularizers~\cite{tartaperlo}. 


%% file: 3_method.tex
\section{Sensitivity-based Regularization for Neurons}
\label{sec:NS}

In this section, we first formulate the sensitivity of a network with respect to the post-synaptic potential of a neuron. Then, we derive a general parameter update rule which relies on the proposed sensitivity term.
As reference scenario, a multi-class classification problem with $C$ labels is considered;  
however, our strategy can be extended to other learning tasks, e.g. regression, in a straightforward way.

\subsection{Preliminaries and Definitions}

\begin{figure}
    \centering
    \includegraphics[width=\columnwidth]{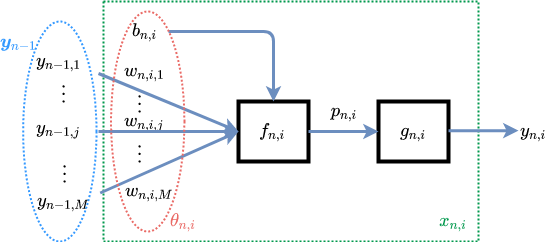}
    \caption{Representation of the neuron $x_{n,i}$ with activation function $g_{n,i}$.}
    \label{fig:nom}
\end{figure}

Let a feed-forward, a-cyclic, multi-layer artificial neural network be composed of $N-1$ hidden layers. We identify with $n = 0$ the input layer and $n = N$ the output layer, other $n$ values indicate the hidden layers. 
For the $i$-th neuron of the $n$-th layer $x_{n,i}$, we define:

\begin{itemize}
  \item ${y}_{n,i}$ as its output,
  \item $\boldsymbol{y}_{n-1}$ as its input vector,
  \item $\theta_{n,i}$ as its own parameters: $\boldsymbol{w}_{n,i}$ the weights and $b_{n,i}$ the bias,
\end{itemize}
\noindent
as illustrated in Fig.~\ref{fig:nom}.
Each neuron has its own \textit{activation function} $g_{n,i}(\cdot)$ to be applied after some affine function $f_{n,i}(\cdot)$ which can be for example convolution or dot product.\\
Hence, the output of a neuron is given by

\begin{equation}
    \label{eq:0}
    \added{y_{n,i} = g_{n,i}[ p_{n,i}]},
\end{equation}
\noindent
\added{where $p_{n,i}$ is the \textit{post-synaptic potential} of $x_{n,i}$ defined as:}
\begin{equation}
    \label{eq:1}
    p_{n,i} = f_{n,i}(\theta_{n,i}, \textbf{y}_{n-1}).
\end{equation}

\subsection{Neuron Sensitivity}


Here we introduce the definition of neuron sensitivity.
We recall that we aim at pruning entire neurons rather than single parameters to achieve structured sparsity.
Let us we assume that our method is applied to a pre-trained network.
To estimate the relevance of neuron $x_{n,i}$ for the task upon which the network was trained, we evaluate the neuron contribution to the network output $\boldsymbol{y}_N$. 
To this end, we first provide an intuition on how small variations of the post-synaptic potential $p_{n,i}$ of neuron $x_{n,i}$ affect the $k$-th output of the network \added{$y_{N,k}$}.
By a Taylor series expansion, for small variations of $p_{n,i}$, let us express the variation of \added{$y_{N,k}$} as

\begin{equation}
    \label{eq:smallperturb}
    \Delta y_{N, k} \approx \Delta p_{n,i} \frac{\partial y_{N, k}}{\partial p_{n, i}}
\end{equation}
\noindent
where $y_{N, k}$ indicates the $k$-th output for the output layer. In the case $\Delta y_{N, k} \rightarrow 0 ,\,\forall k$, for small variations of $p_{n,i}$, $y_{N,k}$ does not change.
Such condition allows to drive the post-synaptic potential $p_{n,i}$ to zero without affecting the network output \added{$y_{N,k}$} (and, for instance, its performance).
Otherwise, if $\Delta y_{N, k}\neq 0$, any variation of $p_{n,i}$  might alter the network output, possibly impairing its performance.

We can now properly quantify the effect of small changes to the network output by defining the \emph{neuron sensitivity}.

\begin{definition}
\label{def:sensitivity}
    The sensitivity of the network output $\boldsymbol{y}_N$ with respect to the post-synaptic potential $p_{n,i}$ of neuron $x_{n,i}$ is:
    \begin{equation}\label{eq:3}
        S_{n,i}(\boldsymbol{y}_N,p_{n,i}) = \frac{1}{C}\sum_{k=1}^{C} \left | \frac{\partial y_{N, k}}{\partial p_{n,i}}\right |
    \end{equation}
    where $\boldsymbol{y}_N \in R^C$ and $S_{n,i} \in [0; +\infty)$.
    Intuitively, the higher $S_{n,i}$, the higher the fluctuation of $\boldsymbol{y}_N$ for small variations of $p_{n,i}$.
\end{definition}
\noindent
Before moving on, we would like to clarify our choice of leveraging the post-synaptic potential $p_{n,i}$ rather than the neuron output $y_{n,i}$ in the equation above.
In order to understand our choice, we re-write \eqref{eq:3} using the chain rule:
\begin{equation}\label{eq:4}
    S_{n,i}(\boldsymbol{y}_N,p_{n,i}) = \frac{1}{C}\sum_{k=1}^{C} \left | \frac{\partial y_{N, k}}{\partial y_{n,i}} \cdot \frac{\partial y_{n, i}}{\partial p_{n,i}}\right |.
\end{equation}
\noindent
Without loss of generality, let us assume $\frac{\partial y_{N,k}}{\partial y_{n,i}}\neq 0$ and $g_{n,i}$ corresponds to the well known ReLU activation function.
Under the hypothesis that $p_{n,i} < 0$, $\frac{\partial y_{n,i}}{\partial p_{n,i}}=0$ for the considered ReLU activation.
Had we written \eqref{eq:3} as a function of the neuron output $y_{n,i}$, the vanishing gradient $\frac{\partial y_{n,i}}{\partial p_{n,i}}=0$ would have prevented us from estimating the neuron sensitivity.
The above consideration applies beyond ReLU to any activation function except for the identity function, for which $y_{n,i} = p_{n,i}$.

\subsection{Bounds on Neuron Sensitivity}
\label{sec:bns}

Here we provide two computationally-efficient bounds to the sensitivity function above that can be practically exploited.
Popular frameworks for DNN training rely on differentiation frameworks such as \emph{autograd}, for automatic variable differentiation along computational graphs. 
Such frameworks take as input some objective function $J$ and automatically compute all the gradients along the computational graph. In order to get $S_{n,i}$ as an outcome from the differentiation engine, we define

\begin{equation}
    S_{n,i}(\boldsymbol{y}_N,p_{n,i}) = \frac{\partial J}{\partial p_{n,i}} 
\end{equation}
\noindent
where $J$ is a proper function. In Appendix~\ref{app:SUP} we show that such function turns to be:
\begin{equation}
    \label{eq:Jexact}
    J=\frac{1}{C} \sum_{k=1}^{C} \int \left | \frac{\partial y_{N, k}}{\partial p_{n,i}}\right | dp_{n,i}
\end{equation}
Therefore, computing the sensitivity in \eqref{eq:3} requires $C$ calls to the differentiation engine.
In the following with some little algebra we derive a lower and upper bound to Def.~\ref{def:sensitivity} that we show to be particularly useful from a computational perspective.\\
Let the objective function to differentiate be
\begin{equation}
    \label{eq:lowR}
    J^{l}=\frac{1}{C}\sum_{k=1}^C y_{N, k}.
\end{equation}
\noindent
The automatic differentiation engine called on $S^l$ will return
\begin{equation}
    \frac{\partial J^{l}}{\partial p_{n, i}} = \frac{1}{C}\sum_{k=1}^C \frac{\partial y_{N, k}}{\partial p_{n,i}}
\end{equation}
\noindent
According to the triangular inequality, a lower bound to the sensitivity in \eqref{eq:3} can be computed as
\begin{equation}
    \label{eq:lowboundS}
    S_{n,i}^l = \frac{1}{C} \left | \sum_{k=1}^{C}  \frac{\partial y_{N, k}}{\partial p_{n,i}}\right | \leq \frac{1}{C}\sum_{k=1}^{C} \left | \frac{\partial y_{N, k}}{\partial p_{n,i}}\right |
\end{equation}
\noindent
$S_{n,i}^l$ can be conveniently evaluated differentiating over \eqref{eq:lowR} (and taking the absolute value) with a single call to the differentiation engine.
As shown in \eqref{eq:lowboundS}, this gives us a lower bound estimation over the neuron sensitivity.\\
In order to estimate an upper bound to $S_{n,i}$, we rewrite \eqref{eq:3} as
\begin{equation}
    \label{eq:senschainrule}
    S_{n,i} = \frac{1}{C} \sum_{k=1}^C \left | \frac{\partial y_{N, k}}{\partial \boldsymbol{y}_{N-1}} \cdot \prod_{l=n+1}^{N-1}\frac{\partial \boldsymbol{y}_{l}}{\partial \boldsymbol{y}_{l-1}} \cdot \boldsymbol{\delta}_{n,i} \frac{\partial y_{n, k}}{\partial p_{n,i}}\right |
\end{equation}
\noindent
However, $\forall k$ we have in common the term
\begin{align}
    \boldsymbol{\Gamma}_{n,i} &= \prod_{l=n+1}^{N-1}\frac{\partial \boldsymbol{y}_{l}}{\partial \boldsymbol{y}_{l-1}} \cdot \boldsymbol{\delta}_{n,i} \frac{\partial y_{n, i}}{\partial p_{n,i}}\nonumber\\
    &\leq \prod_{l=n+1}^{N-1} \left| \frac{\partial \boldsymbol{y}_{l}}{\partial \boldsymbol{y}_{l-1}}\right| \cdot \boldsymbol{\delta}_{n,i} \left|\frac{\partial y_{n, i}}{\partial p_{n,i}}\right| = \boldsymbol{\Gamma}_{n,i}^u
\end{align}
\noindent
where $\delta_{n,i}$ is a one-hot vector selecting the $i$-th neuron at the $n$-th layer and $|\cdot|$ is an element-wise operator.
Hence, we rewrite \eqref{eq:senschainrule} as
\begin{equation}
	\label{eq:upboundS}
	 S^{u}_{n,i} = \frac{1}{C} \left(\sum_{k=1}^C \left| \frac{\partial y_{N, k}}{\partial \boldsymbol{y}_{N-1}} \right|\right) \cdot \boldsymbol{\Gamma}_{n,i}^u \geq S_{n,i}.
\end{equation}
\noindent
Thus, we have shown that $S_{n,i}^u$ is an upper bound to the sensitivity in \eqref{eq:3}. 
\noindent
Upper and lower bounds are here obtained for two main reasons: computational efficiency and relaxing/tightening conditions on the sensitivity itself. We will see in Sec.~\ref{sec:prel_exp} a typical population distribution of the sensitivities on a pre-trained network, comparing \eqref{eq:3}, \eqref{eq:lowboundS} and \eqref{eq:upboundS}.
In the following, we exploit the formulation of the the Sensitivity function \eqref{fig:nom} and its two bounds \eqref{eq:lowboundS}, \eqref{eq:upboundS} to define a parameter update rule.

\subsection{Parameters Update Rule}
Now we show how the proposed sensitivity definition can be exploited to promote neuron sparsification.
As hinted before, if the sensitivity $S_{n,i}$ of neuron $x_{n,i}$ is small, i.e $S_{n,i} \rightarrow 0$, then neuron $x_{n,i}$ yields a small contribution to the $i$-th network output $y_{N,i}$ and its parameters may be moved towards zero with little perturbation to the network output.
To this end, we define the \textit{insensitivity} function $\overline{S}_{n, i}$ as
\begin{equation}
    \label{eq:insensitivity}
    \overline{S}_{n, i} = \max \{ 0, 1-S_{n,i} \} =  \left(1-S_{n,i}\right) \cdot \Theta \left(1-S_{n,i}\right)
\end{equation}
\noindent
where $\Theta(\cdot)$ is the one-step function. The higher the insensitivity of neuron $x_{n,i}$ (i.e., $\overline{S}_{n, i} \rightarrow 1$ or equivalently $S_{n, i} \rightarrow 0$), the less the neuron affects the network output.
Therefore, if  $\overline{S}_{n, i} \rightarrow 1$, then neuron $x_{n,i}$ contributes little to the network output and its parameters $w_{n,i,j}$ can be driven towards zero without significantly perturbing the network output.
Using the insensitivity definition in \eqref{eq:insensitivity}, we propose the following update rule:

\begin{equation}
    \label{eq:updaterule}
    w_{n,i,j}^{t+1} = w_{n,i,j}^{t} - \eta \frac{\partial L}{w_{n,i,j}^{t}} - \lambda w_{n,i,j}^{t} \overline{S}_{n, i}
\end{equation}
\noindent
where 
\begin{itemize}
    \item the first contribution term is the classical minimization of a loss function $L$
    , ensuring that the network still solves the target task, e.g. classification;
    \item the second one represents a penalty applied to the parameter $w_{n,i,j}$ belonging to the neuron $x_{n,i}$ which is proportional to the insensitivity of the output to its variations.
\end{itemize}
\noindent
Finally, since
\begin{equation}
    \label{eq:derivisw}
    \frac{\partial p_{n,i}}{\partial y_{n-1,j}} = w_{n,i,j}
\end{equation}
we rewrite \eqref{eq:updaterule} as
\begin{equation}
    \label{eq:updaterule2}
    w_{n,i,j}^{t+1} = w_{n,i,j}^{t} - \eta \frac{\partial L}{w_{n,i,j}^{t}} - \lambda \dot{S}_{n, i, j}
\end{equation}
where
\begin{equation}
    \label{eq:Sdot}
    \dot{S}_{n, i, j} = \left[ w_{n,i,j}-\frac{\mbox{sign}(w_{n,i,j})}{C} \sum_{k=1}^C \left| \frac{\partial y_{N, k}}{\partial y_{n-1, j}} \right| \right] \cdot \Theta \left(1-S_{n,i}\right)
\end{equation}
A step-by-step derivation is provided in Appendix~\ref{app:17deriv}.
From \eqref{eq:Sdot} we can better understand the effect of the proposed penalty term: as expected by our discussion above, $\dot{S}_{n, i, j}$ is inversely proportional to the impact on the output for variations of the input for the neuron $x_{n,i}$. 

\subsection{Local neuron sensitivity-based regularization}

We propose now an approximate formulation of the sensitivity function in \eqref{eq:3} based only on the post-synaptic potential and output of a neuron that we will refer to as the \emph{local} sensitivity.
Let us recall that for each neuron $x_{n,i}$ the sensitivity provided by Definition~\ref{def:sensitivity} measures the overall impact of a given neuron $x_{n,i}$ on the network output taking into account all the following neurons involved in the computation.

\begin{definition}
    The \emph{local} neuron sensitivity of the output $y_{n, i}$ with respect to the post-synaptic potential $p_{n,i}$ of the neuron $x_{n,i}$ is defined as:
    \begin{equation}\label{eq:5}
        \tilde{S}_{n,i} = \left | \frac{\partial y_{n,i}}{\partial p_{n,i}}\right |
    \end{equation}
\end{definition}
In the case of ReLU-activated networks, it simply reads
\begin{equation}
    \tilde{S}_{n,i} = \Theta(p_{n,i})
\end{equation}
Under this setting, the update rule \eqref{eq:updaterule2} simplifies to
\begin{equation}
    \label{eq:updaterulelocal}
    w_{n,i,j}^{t+1} = w_{n,i,j}^{t} - \eta \frac{\partial L}{w_{n,i,j}^{t}} - \lambda w_{n,i,j}^{t}\Theta(-p_{n,i}),
\end{equation}
i.e., the penalty is applied only in case the neuron stays off.
While local sensitivity is a looser approximation of \eqref{def:sensitivity}, it is far less complex to compute especially for ReLU-activated neurons.

%% file: 4_training_new.tex
\section{The SeReNe procedure}
\label{sec:training}

\begin{figure}
	\centering
			\centering
			\includegraphics[width=0.78\columnwidth]{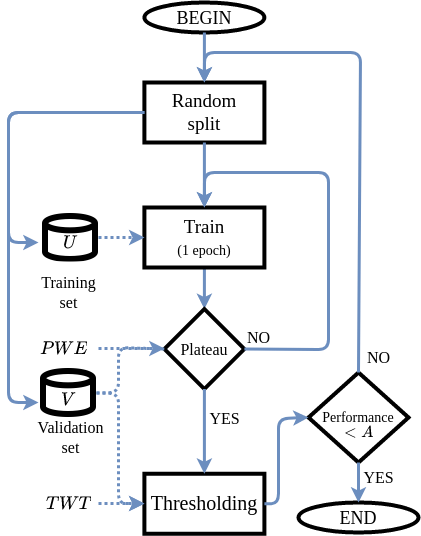}
			\caption{\added{High-level view of the SeReNe procedure.}}
			\label{fig:high-levelserene}
\end{figure}

This section introduces a practical procedure to prune neurons from a neural network $\mathcal{N}$ leveraging the sensitivity-based regularizer introduced above.
Let us assume $\mathcal{N}$ has been preliminary trained at some task over the dataset $D$ achieving performance (e.g., classification accuracy) $A$. 
We do not put any constraint over the actual training method, training set or network architecture.
Alg.~\ref{code:main} summarizes the procedure in pseudo-code.
In a nutshell, the procedure consists in iteratively looping over the \textit{Regularization} and \textit{Thresholding} procedures.
At the beginning of the loop, dataset $D$ is split into disjoint subset $V$ (used for validation purposes) and $U$ (to update the network).
At line 5, the regularization procedure (summarized in Alg. \ref{code:regularize}) trains $\mathcal{N}$ over $D$ according to \eqref{eq:updaterule} driving towards zero parameters of neurons with low sensitivity.
The loop ends if the performance of the regularized network falls below threshold $A$.
Otherwise, the thresholding procedure sets to zero parameters below threshold $T$ and prunes neurons such that all parameters are equal to zero.
The output of the procedure is the pruned network, i.e. with fewer neurons, $\mathcal{N}^{\star}$.
The Regularization and Thresholding procedures are detailed in the following. \added{A graphical high-level representation of SeReNe is also displayed in Fig.~\ref{fig:high-levelserene}.}

\begin{algorithm}[!ht]
    \caption{The SeReNe procedure}\label{code:main}
    \hspace*{20pt}\textbf{Input: }\textit{Trained network $\mathcal{N}$, dataset D,\\\hspace*{20pt}Target performance A, PWE, TWT\\}
    \hspace*{20pt}\textbf{Output: }\textit{Pruned network $\mathcal{N}^{\star}$}
    \begin{algorithmic}[1]
        \Procedure{SeReNe}{$\mathcal{N}, D, A, PWE, TWT$}
        \State $\mathcal{N^{\star}} \gets \mathcal{N}$
        \While {true}
        \State $U, V \gets \textsc{RandomSplit}(D)$
        \State $\mathcal{N} \gets \textsc{Regularization}(\mathcal{N}, U, V, PWE)$
        \If {$\textsc{Performance}(\mathcal{N}, V) < A$}
            \State break
        \EndIf
        \State $\mathcal{N}^{\star} \gets \mathcal{N}$
        \State $\mathcal{N} \gets \textsc{Thresholding}(\mathcal{N}, V, TWT)$
        \EndWhile
        \Return{$\mathcal{N}^{\star}$}
        \EndProcedure
    \end{algorithmic}
\end{algorithm}

\subsection{Regularization}
\label{sec:Regularization}

This procedure takes in input a network $\mathcal{N}$ and returns a regularized network according to the update rule \eqref{eq:updaterule}.
Namely, the procedure iteratively trains $\mathcal{N}$ on $U$ and validates it on $V$ for multiple epochs.
Let $\mathcal{N}^r$ represent the \textit{best} regularized network found at a given time according to the loss function.
For each iteration, the procedure operates as follows.
First (line 5), $\mathcal{N}$ is trained for one epoch over $U$: the results is a regularized network according to \eqref{eq:updaterule}.
Second (line 6), this network is validated on $V$.
If the loss is lower than the loss of $\mathcal{N}^r$ over $V$, then $\mathcal{N}$ takes the place of $\mathcal{N}^r$ (line 7).
If $\mathcal{N}^r$ is not updated for $\textrm{PWE}$ (\emph{Plateau Waiting Epochs}) epochs, we assume we have reached a performance plateau.
In this case, the procedure ends and returns the sensitivity-regularized network $\mathcal{N}^r$.

\begin{algorithm}
    \caption{The regularization procedure}\label{code:regularize}
    \hspace*{20pt}\textbf{Input: }\textit{Model $\mathcal{N}$, data sets V and U, PWE}\\
    \hspace*{20pt}\textbf{Output: }\textit{The sensitivity-regularized network $\mathcal{N}^r$}
    \begin{algorithmic}[1]
        \Procedure{Regularization}{$\mathcal{N}, U, V, PWE$}
        \State $\mathcal{N}^r \gets \mathcal{N}  $\Comment{$\mathcal{N}^r$ is \textit{best} regularized network on $V$}
        \State $epochs \gets 0$
        \While {$epochs < PWE$}
            \State $\mathcal{N} \gets \textsc{Train}(\mathcal{N}, U)$\Comment{1 train epoch on $U$}
            \State $epochs++$
            \If{$\textsc{Loss}(\mathcal{N},V) < \textsc{Loss}(\mathcal{N}^r,V)$}
                \State{$\mathcal{N}^r \gets \mathcal{N}$}
                \State $epochs \gets 0$
            \EndIf
        \EndWhile
        \Return{$\mathcal{N}^r$}
        \EndProcedure
    \end{algorithmic}
\end{algorithm}

\subsection{Thresholding}
\label{sec:Threshold}

The \emph{thresholding} procedure is where the parameters of neurons with low sensitivity are thresholded to zero.
Namely, parameters whose absolute value is below threshold $T$ are pruned as 

\begin{equation}
    w_{n,i,j} = \left \{ 
    \begin{array}{ll}
       w_{n,i,j} & \left | w_{n,i,j} \right | > T\\
       0    & otherwise.
    \end{array}
    \right .
\end{equation}
\noindent
The pruning threshold $T$ is selected so that the performance (or, in other words, the loss on $V$) worsens at most of a relative value we call \emph{thresholding worsening tolerance} ($TWT$) we provide as hyper-parameter.\\
We expect the loss function to be locally a smooth, monotone function of $T$, for small values of $T$.
The threshold $T$ can be found using linear search-based heuristics. We can however reduce this using a bisection approach, converging to the optimal $T$ value in log-time steps. 

Because of the stochasticity introduced by mini-batch based optimizers, parameters pruned during a thresholding iteration may be reintroduced by the following regularization iteration.
In order to overcome this effect, we enforce that pruned parameters can no longer be updated during the following regularizations (we term this behavior as \emph{parameter pinning}).
To this end, the update rule \eqref{eq:updaterule} is modified as follows:
\begin{equation}
    w_{n,i,j}^{t+1} = \left\{ 
    \begin{array}{ll}
        w_{n,i,j}^{t} - \eta \frac{\partial L}{w_{n,i,j}^{t}} - \lambda w_{n,i,j}^{t} \overline{S}_{n, i} & w_{n,i,j}^{t} \neq 0\\
        0 & w_{n,i,j}^{t} = 0
    \end{array}
    \right .
    \label{eq:updatewmem}
\end{equation}
\noindent
We have noticed that without parameter pinning, the compression of the network may remain low because the noisy gradient estimates in a mini-batch that keep reintroducing previously pruned  parameters.
On the contrary, by adding \eqref{eq:updatewmem} a lower number of epochs are sufficient to achieve much higher compression.

%% file: 5_results.tex
\section{Results}
\label{sec::results}

\input{tables/lenet300_1.65_tab}
\input{tables/lenet300_1.95_tab}

\added{

In this section we experiment with our proposed neuron pruning method comparing the four sensitivity formulations we introduced in the previous section:
\begin{itemize}
    \item SeReNe (exact) - the exact formulation in~\eqref{eq:4};
    \item SeReNe (LB) - the lower bound in~\eqref{eq:lowboundS};
    \item SeReNe (UB) - the upper bound in~\eqref{eq:upboundS};
    \item SeReNe (local) - the local version in~\eqref{eq:5};
    \item $\ell_2$~+~pruning - is a baseline reference where we replace our sensitivity-based regularization term with a standard $\ell_2$ term (all the rest of the framework is identical).
\end{itemize}
\noindent
We experiment over different combinations of architectures and datasets commonly used as benchmarks in the relevant literature:

\begin{itemize}
    \item LeNet-300 on MNIST (Table~\ref{tab:lenet300_16} and Table~\ref{tab:lenet300_19}),
    \item LeNet-5 on MNIST (Table~\ref{tab:lenet5}),
    \item LeNet-5 on Fashion-MNIST (Table~\ref{tab:lenet5fashion}),
    \item VGG-16 on CIFAR-10 (Table~\ref{tab:VGG1} and Table~\ref{tab:VGG2}),
    \item ResNet-32 on CIFAR-10 (Table~\ref{tab:resnet32}),
    \item AlexNet on CIFAR-100 (Table~\ref{tab:Alexnet}),
    \item ResNet-101 on ImageNet (Table~\ref{tab:resnet101}).
\end{itemize}
\noindent
Notice that the VGG-16, AlexNet and ResNet-32 architectures are modified to fit the target classification task (CIFAR-10 and CIFAR-100).
The validation set ($V$) size for all experiments is $10\%$ of the training set.
\\
The pruning performance is evaluated according to multiple metrics.
\begin{itemize}
    \item The \textit{compression ratio} as the ratio between the number of parameters in the original network and the number of remaining parameters after pruning (the higher the better).
    \item The number of remaining neurons (or filters for convolutional layers) after pruning. 
    \item The size of the networks when stored on disk in the popular ONNX format~\cite{bai2019} (\texttt{.onnx} column).
    ONNX files are then lossless compressed using the Lempel–Ziv–Markov algorithm (LZMA)~\cite{pavlov2007lzma} (\texttt{.7z} column). 
\end{itemize}
\noindent
In our experiments, we compare with all available references for each combination of architecture and dataset. For this reason, the reference set may vary from experiment to experiment.
Our algorithms are implemented in Python, using PyTorch~1.5, and simulations are run on a RTX2080~NVIDIA~GPU with 8GB of memory.\footnote{The source code will be made available upon acceptance of the article.}
}

\subsection{Preliminary experiment}
\label{sec:prel_exp}

\begin{figure}
    \includegraphics[width=\columnwidth]{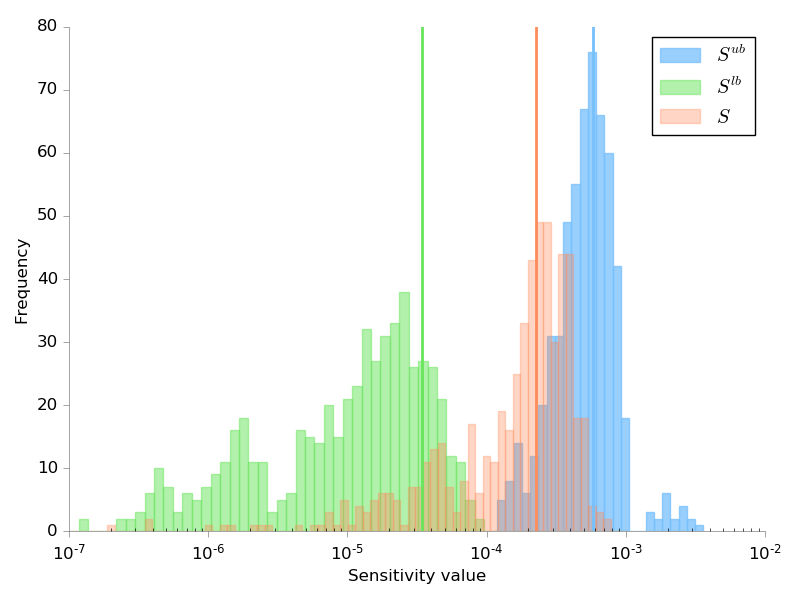}
    \caption{Population of sensitivities $S$ and relative lower $S^l$ and upper $S^u$ bounds for a LeNet-5 architecture pre-trained on MNIST. Vertical bars indicate relative mean values.}
    \label{fig:sensitivities}
\end{figure}
To start with, we plot the sensitivity distribution for a LeNet-5 network trained on the MNIST dataset (SGD with learning rate $\eta = 0.1$ weight-decay $10^{-4}$). This network will also be used as baseline in Sec.~\ref{sec:ln5mnist}. Fig.~\ref{fig:sensitivities} shows SeReNe (exact) (red), SeReNe (LB) (green) and SeReNe (UB) (blue); the vertical mars represent the mean values. As expected, SeReNe (LB) and SeReNe (UB) under estimate and over estimate SeReNe (exact), respectively.
Interestingly, SeReNe (UB) sensitivity values lie in the range $[10^{-4};10^{-2}]$ while both for SeReNe (exact) and SeReNE (LB) show a longer trail towards smaller figures, whereas all distributions look similar.
In the following, we will experimentally evaluate the three sensitivity formulations in terms of pruning effectiveness.

\subsection{LeNet300 on MNIST}
\label{sec:ln300}

\begin{figure}[!b]
    \begin{center}
        \begin{subfigure}{\columnwidth}
                \centering
                \advance\leftskip-0.3cm
                \includegraphics[width=0.78\columnwidth, height=0.3\columnwidth]{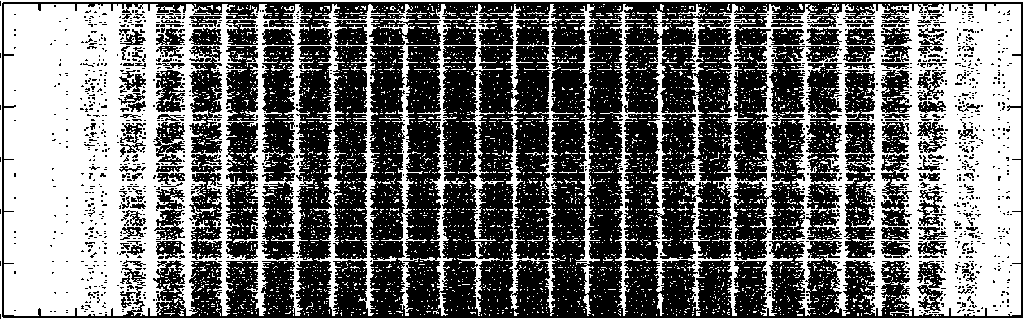}
        \end{subfigure}
    \end{center}
    \vskip\baselineskip
    \begin{center}
        \begin{subfigure}{\columnwidth}
                \centering
                \includegraphics[angle=-90,width=\columnwidth]{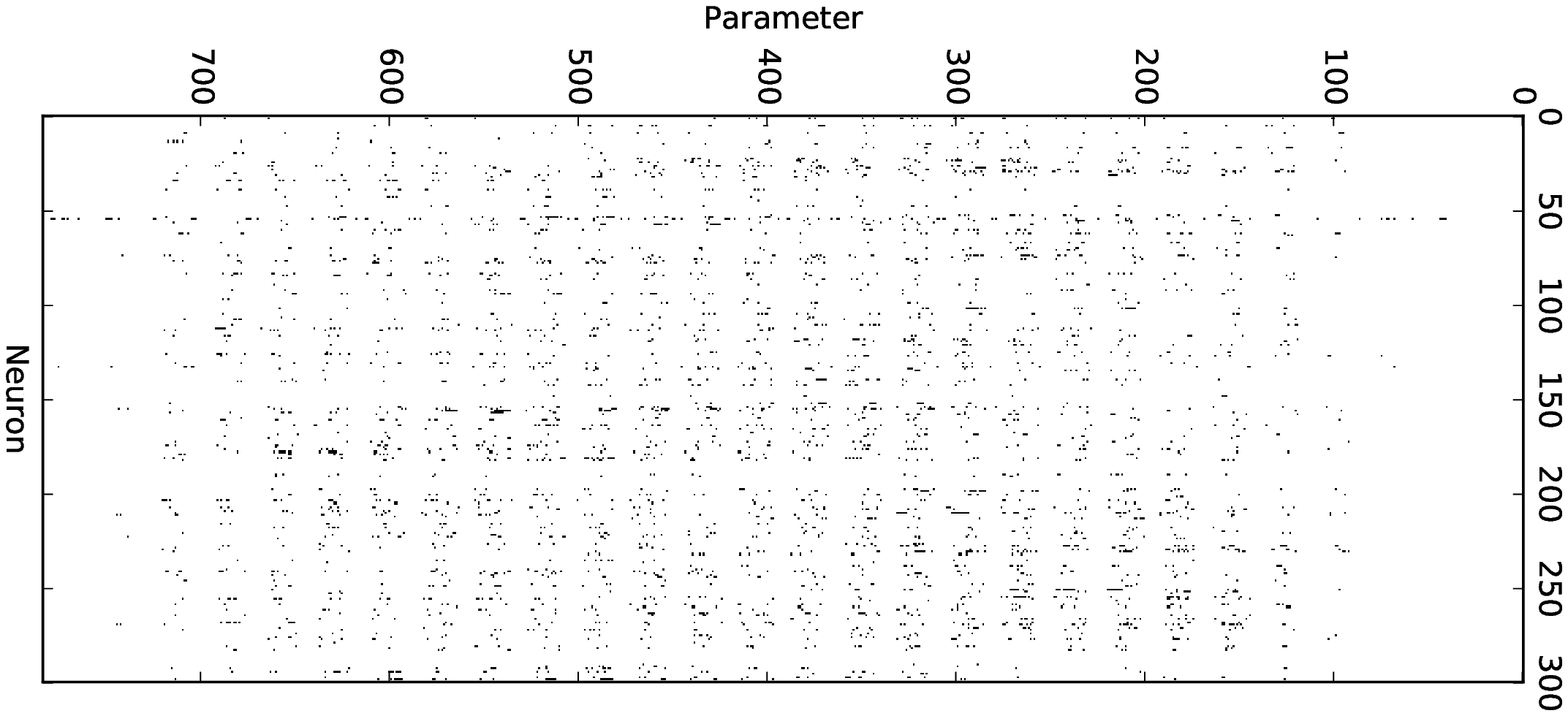}
                \label{fig:fc1lenet300}
        \end{subfigure}
    \end{center}
    \caption{Parameters distribution in FC1 of LeNet-300 trained on MNIST from Han~\emph{et~al.}~\cite{han:deep_compression} (top) and he proposed SeReNe (bottom). In black the remaining parameters.
    }
    \label{fig:TOF}
\end{figure}

As a first experiment, we prune a LeNet-300 architecture, which consists of three fully-connected layers with 300, 100 and 10 neurons, respectively trained over the MNIST dataset.
We pre-trained LeNet-300 via SGD with learning rate $\eta = 0.1$ and $PWE=20$ epochs with $\lambda=10^{-5}$, $TWT=0.3$ for SeReNe (exact), SeReNe~(LB) SeReNe~(UB) and $\lambda=10^{-5}$, $TWT=1$ for SeReNe~(local).
The related literature reports mainly i) results for classification errors around $1.65\%$ (Table~\ref{tab:lenet300_16}) and ii) results for errors in the order of $1.95\%$ (Table~\ref{tab:lenet300_19}).
For this reason, we trained for about 1k epochs to achieve $1.95\%$ error rate and for additional 2k epochs to score a $1.65\%$ error rate. 
\\
SeReNe outperforms the other methods leads both in terms of compression ratio and number of pruned neurons.
SeReNe (exact) achieves a compression ratio of 42.55$\times$ and the number of remaining neurons in the hidden layers drops from 300 to 159 and from 100 to 75 respectively.
\input{tables/lenet5_mnist_tab}
\input{tables/lenet5_fmnist_tab}
SeReNe (LB) enjoys comparable performance with respect to SeReNe (exact) despite lower computational cost (see below).
For the \.95\% error band, SeReNe (LB) performs is more effective at pruning parameters than SeReNe (exact), allowing lower error. 
Serene (LB) prunes more parameters than SeReNe (UB), we hypothesize because \eqref{eq:upboundS} overstimates the sensitivity of the parameters and prevents them to be pruned.
On the other side, SeReNe (LB) underestimates the sensitivity, however small $\lambda$ values sets this off.
SeReNe (local) prunes less parameters than the other SeReNe formulations as it relies on a locally computed sensitivity formulation despite lower complexity.
Concerning training time (second column from the right), SeReNe (local) is fastest and introduces very little computational overhead, SeReNe (UB) and SeReNe (LB) have comparable training times and the slowest is the SeReNe (exact), approximately 2.7x slower than its boundaries.
In the light of the good tradeoff between ability to prune neurons, error rate and training time of SeReNe (LB), in the following we will restrict our experiments to this sensitivity formulation.

\noindent
Fig.~\ref{fig:TOF} (bottom) shows the location of the parameters not pruned by SeReNe (exact) in LeNet300 first fully-connected layer (black dots).
For comparison, we report the equivalent image from Fig.~4 of \cite{han:deep_compression} (top).
Our method yields completely blank columns in the matrix that can be represented in memory as uninterrupted sequences of zeroes.
When stored on disk, LZMA compression (.7z column) is particularly effective at encoding long sequences of the same symbol, which explains the 10x compression rate it achieves (from 538 to 46 kB) over the .onnx file.

Finally, we perform an ablation study to assess the impact of a simpler $\ell_2$-only regularization, i.e. classical weight decay, in place of our sensitivity-based regularizer. 
Towards this end, we retrain LeNet-300 with $\lambda=0$ and a weight-decay set to $10^{-4}$ in its place (line $\ell_2$+pruning in the tables above).
We point out in \eqref{eq:updaterule} that the sensitivity can be interpreted as a weighting factor for the $\ell_2$-regularization.
Using weight-decay is equivalent to assuming all the parameters have the same sensitivity.
For this experiment, we used $\eta=0.1$, $PWE=5$ and $TWT=0$ ($TWT>0$ significantly and uncontrollably worsens the performance).
Table~\ref{tab:lenet300_16} shows that such method is less effective at pruning neurons than SeReNe (LB), which removes 15\%  more neurons.
Similar conclusions can be drawn also if higher error is tolerated, as in Table~\ref{tab:lenet300_19}. The $\ell_2$+pruning has been performed for comparison in all following experiments in the paper yielding the same results.


\subsection{LeNet5 on MNIST}
\label{sec:ln5mnist}


Next, we repeat the previous experiment over the LeNet-5~\cite{lecun1998gradient} architecture, preliminarily trained as for the LeNet-300 above, yet with SGD with learning rate $\eta = 0.1$ and $PWE=20$ epochs.
We experiment with SeReNe (LB) with parameters  ($\lambda=10^{-4}$, $TWT=1.45$).
For this architecture, our method requires about 500 epochs to achieve the same error range as other state of the art references.
According to Table~\ref{tab:lenet5}, SeReNe~(LB) approaches the classification accuracy of its competitors outperforms the considered references in terms of compression ratio and pruned neurons.\\
In this case, the benefits coming from the structured sparsity are evident: the uncompressed network storage footprint decreases from 1686~kB to 208~kB (-90\%), which after lossless compression further decreases to 19~kB with a 0.12\% performance drop only.  

\input{tables/vgg1_cifar10_tab}
\input{tables/vgg2_cifar10_tab}
\input{tables/resnet32_cifar10}

\subsection{LeNet5 on Fashion-MNIST}


Then, we experiment with the same LeNet-5 architecture on the Fashion-MNIST~\cite{FMNIST} dataset.
Fashion-MNIST has the same size of the MNIST dataset, yet it contains natural images of dresses, shoes, etc. and so it is harder to classify than MNIST since the images are not sparse as MNIST digits.
In this experiment we used SGD with learning rate $\eta = 0.1$ and $PWE=20$ epochs. For SeReNe~(LB) we used $\lambda=10^{-5}$ and $TWT=1$ for about 2k epochs.\\
Unsurprisingly, the average compression ratio is lower than for MNIST: since the classification problem is much harder than MNIST (Sec.~\ref{sec:ln5mnist}), more complexity is required and SeReNe, in order not to degrade the Top-1 performance, is not pruning as much as it did for the MNIST experiment. Most importantly, the SeReNe~(LB) compressed network is 46~kB only, despite the higher number of pruned parameters.

\subsection{VGG on CIFAR-10.}

\added{
Next, we experiment with two popular implementations of the VGG architecture~\cite{simonyan:vgg}.
We recall that VGG consists in 13 convolutional layers arranged in 5 groups of, respectively, 2, 2, 3, 3, 3 layers, with 64, 128, 256, 512, 512 filters per layer respectively.
\textit{VGG-1} is a VGG implementation popular in CIFAR-10 experiments that includes only one fully-connected layer as output layer and is pre-trained on ImageNet~\footnote{https://github.com/kuangliu/pytorch-cifar}.
\textit{VGG-2}~\cite{molchanov:var_dropout} is similar to VGG-1 but includes one hidden fully connected layer with 512 neurons before the output layer.
We experiment over the CIFAR-10 dataset, which consists of 50k $32 \times 32$, RGB images for training and 10k for testing, distributed in 10 classes.
For both VGG-1 and VGG-2 we have used SGD with learning rate $\eta = 0.01$ and $PWE=20$ epochs. For the SeReNe (LB), we used $\lambda=10^{-6}$ and $TWT=1.5$.
Both architectures were pruned for approximately 1k epochs and Tables~\ref{tab:VGG1} and ~\ref{tab:VGG2} detail the pruned topologies.
For each architecture, we detail the number of surviving filters (convolutional layers) or neurons (fully connected layers) for each layer within square brackets. 
The tables show that SeReNe introduces a significantly structured sparsity for both VGG-1 and VGG-2 and outperforms Sparse-VD~\cite{molchanov:var_dropout} in terms of compression ratio.
We are able to prune a significant number of filters also in the convolutional layers; as an example, the 3 layers in block Conv4 are reduced to [382]-[93]-[136] for VGG-1 and [498]-[433]-[65] for VGG-2.
That has a positive impact on the networks footprint.
VGG-1 memory footprint drops from 57.57~MB to 11.56~MB for the pruned network, while the \textit{7zip} compressed representation is 0.97~MB only.
For VGG-2, the memory foot print drops from 58.61~MB to 29.41~MB, while the compressed file representation amounts to 2.47~MB.
}

\input{tables/alexnet_cifar100_tab}
\input{tables/resnet101_imagenet_tab}


\subsection{ResNet-32 on CIFAR-10}

We then evaluate SeReNe over the ResNet-32 architecture~\cite{he2016deep} trained on the CIFAR-10 dataset using SGD with learning rate $\eta=0.001$, momentum $0.9$, $\lambda=10^{-5}$, $TWT=0$ and $PWE=10$.
Table~\ref{tab:resnet32} shows the resulting architecture.
Due to the number of layers, we represent the network architecture in five different blocks: the first correspond to the first convolutional layer that takes in input the original input image, the last represent the fully-connected output layer. The other three blocks in the middle represent the rest of the network, based on the number of output channels of each layer: \emph{block1} contains all the layers with an output of 16 channels, \emph{block2} contains all the layers with an output of 32 channels and \emph{block3} collects the layers with an output of 64 channels.
ResNet is an already optimized architecture and so it is more challenging to prune compared to, e.g, VGG.
Nevertheless, SeReNe is still able to prune about 40\% of the neurons and 70 \% of the parameters over the original ResNet-32.
This is reflected on the size of the network, which drops from 1.84 MB (1.63 MB compressed) to 0.87 MB (0.57MB compressed).

\subsection{AlexNet on CIFAR-100}

Next, we up-scale in the output dimensionality of the learning problem, i.e. in the number of classes $C$, testing the proposed method on an AlexNet-like network over the CIFAR-100 dataset.
Such dataset consists of $32\times32$ RGB images divided in 100 classes (50k training images, 10k test images). In this experiment we use SGD with learning rate $\eta = 0.1$ and $PWE=20$ epochs.
Concerning SeReNe (LB), we used $\lambda=10^{-5}$ and $TWT=1.5$
and the pruning process lasted 300 epochs.\\
Table~\ref{tab:Alexnet} shows compression ratios in excess of 179x, whereas the network size drops from 92.31~MB to 43.80~MB and further to 2.47~MB after compression.\\
With respect to CIFAR-10, we hypothesize that the larger number of target classes to discriminate prevents pruning neurons in the convolutional layers, yet it allows to prune a significant number of neurons from the hidden fully connected layers. 
Contrarily from the previous experiments, the top-5 and the top-1 errors \emph{improve} with respect to the baseline.

\subsection{ResNet-101 on ImageNet}

As a last experiment, we test SeReNe on ResNet-101 trained over ImageNet (ILSVRC-2012), using the pre-trained network provided by the torchvision library.\footnote{\url{https://pytorch.org/docs/stable/torchvision/models.html}}\\
Due to the long training time, we employed a batch-wise heuristic such that, instead of waiting for a performance plateau, the pruning step is taken every time a fifth of the train set (around 7.9k iterations) has been processed. We trained the network using SGD with a learning rate $\eta = 0.001$ and momentum $0.9$; for SeReNe (LB) we used $\lambda=10^{-6}$ and $TWT=0$.\\
Table~\ref{tab:resnet101} shows the result of the pruning procedure with the layers grouped in blocks similarly as for the ResNet-32 experiment. 
Despite the complexity of the classification problem (1000 classes) that makes challenging pruning entire neurons,
we prune around 86\% of the parameters and obtain a network that is smaller in size, especially when compressed, going from 156.67 MB to only 27.84 MB. 

\added{
\subsection{Experiments on mobile devices}
\begin{table}
\renewcommand{\arraystretch}{1.3}
\caption{Inference measures on Huawei P20.}
\label{tab:infer}
\centering

    \begin{tabular}{c c c }
    \hline
    Architecture    & Approach      & Inference time [ms]   \\
    \hline
    ResNet-32
                    &Baseline       & $32.12\pm 3.62$       \\
                    &SeReNe (LB)    & $\mathbf{24.83\pm 3.59}$       \\
    VGG-16 (VGG-1)
                    &Baseline       & $204.21\pm 6.05$      \\
                    &SeReNe (LB)    & $\mathbf{98.67\pm 8.71}$      \\
    AlexNet
                    &Baseline       & $131.41 \pm 11.04$    \\
                    &SeReNe (LB)    & $\mathbf{75.27 \pm 8.70}$     \\
    \hline
    \end{tabular}
\end{table}

As a last experiment, we benchmark some of the architectures pruned with SeReNe on an a Huawei P20 smartphone equipped with 4x2.36 GHz Cortex-A73 + 4x1.84GHz Cortex-A53 processors and 4GB RAM, running Android~8.1 ``Oreo''.
Table~\ref{tab:infer} shows the the inference time for ResNet-32, VGG-16 and AlexNet (all figures are obtained averaging 1,000 inferences on the device).
SeReNe-pruned architectures show consistently lower inference time in the light of the fewer neurons in the pruned network, with a top speedup for VGG-16 in excess of a 2x factor.
These results do not account for strategies commonly employed to boost inference speed, like parameters quantization or custom libraries for sparse tensors processing.
We hypothesize that such strategies, being orthogonal to neuron pruning, would further boost inference time.
}

%% file: tables/lenet300_1.65_tab.tex
\begin{table*}[tb]
	\renewcommand{\arraystretch}{1.3}
	\caption{LeNet-300 trained on MNIST (1.65\% error rate).}
	\label{tab:lenet300_16}
	\centering
		\begin{tabular}{c c c c c c c c c c c}
			\hline
			\multirow{2}{*}{Approach} & \multicolumn{3}{c}{Remaining parameters (\%)} & Compr. & Remaining & \multicolumn{3}{c}{Network size [kB]} & Training time & Top-1\\
			& FC1 & FC2 & FC3 & ratio & neurons & \texttt{.onnx} & & \texttt{.7z} & (s/epoch) & (\%)\\
			\hline
			Baseline                                                   &100         & 100        & 100         & 1x         &[300]-[100]-[10]             &1043&$\rightarrow$&933      &{\bf 3.65}&{\bf 1.44} \\
			Han \textit{et al.} \cite{han:deep_compression}            &8         &9         &26         &12.2x       & -                           &&-&                  &-&1.6        \\
			Tartaglione \textit{et al.} \cite{tartaglione:sensitivity} &2.25      &11.93     &69.3       &27.87x      & -                           &&-&                   &-&1.65       \\
			$\ell_2$+pruning                                           &2.44      &15.76     &68.50      &23.26x      &[212]-[82]-[10]              &723&$\rightarrow$&64   &3.65   &1.66\\
			SeReNe (exact)                                                 &{\bf 1.42}&{\bf 9.54}&60.9       &{\bf 42.55x}&{\bf [159]}-{\bf [75]}-[10]  &{\bf 538}&$\rightarrow$&{\bf 46} &13.25&1.64        \\
			SeReNe (UB)                                              &22.45     &60.81     &87.75      &3.71x       &[295]-[92]-[10]              &1016&$\rightarrow$&324      &5.13&1.67       \\
			SeReNe (LB)                                              &1.51      &10.05     &{\bf 60.53}&39.79x      &[164]-[78]-[10]              &557&$\rightarrow$&55       &4.88&1.65       \\
			SeReNe (local)                                           &3.85      &32.53     &73.49      &13.81x      &[251]-[86]-[10]              &859&$\rightarrow$&119      &3.83&1.64       \\
			\hline
	\end{tabular}
\end{table*}

%% file: tables/lenet300_1.95_tab.tex
\begin{table*}[tb]
	\renewcommand{\arraystretch}{1.3}
	\caption{LeNet-300 trained on MNIST (1.95\% error rate).}
	\label{tab:lenet300_19}
	\centering
		\begin{tabular}{c c c c c c c c c c c}
			\hline
			\multirow{2}{*}{Approach} & \multicolumn{3}{c}{Remaining parameters (\%)} & Compr. & Remaining & \multicolumn{3}{c}{Network size [kB]} & Training time & Top-1\\
			& FC1 & FC2 & FC3 & ratio & neurons & \texttt{.onnx} & & \texttt{.7z} & (s/epoch) & (\%)\\
			\hline
			Baseline                                                   &100         & 100        & 100         & 1x         &[300]-[100]-[10]             &1043&$\rightarrow$&933      &{\bf 3.65}&{\bf 1.44} \\
			Sparse~VD~\cite{molchanov:var_dropout}                     &1.1       &2.7       & 38        &68x         & -                           &&-&                  &-&1.92\\
			SWS~\cite{ullrich:weight_sharing}                          &-         &-         & -         &23x         & -                           &&-&                  &-&1.94\\
			Tartaglione~\textit{et al.}~\cite{tartaglione:sensitivity} &0.93      &{\bf 1.12}& 5.9       &{\bf 103x}  & -                           &&-&                  &-&1.95\\
			DNS~\cite{guo:surgery}                                     &1.8       &1.8       &{\bf 5.5}  &56x         & -                           &&-&                  &-&1.99\\
			$\ell_2$+pruning                                           &1.22      &8.77      &61.10      &41.95x      &[167]-[76]-[10]              &566&$\rightarrow$&42   &3.65   &1.97\\
			SeReNe (exact)                                                  &0.76      &5.85      &49.77      &66.28x      &[148]-{\bf [70]}-[10]        &498&$\rightarrow$&38       &13.25&1.93       \\
			SeReNe (UB)                                              &13.67     &50.76     &84.47      &5.47x       &[293]-[91]-[10]              &1008&$\rightarrow$&240      &5.13&1.95       \\
			SeReNe (LB)                                              &{\bf 0.75}&5.79      &49.3       &66.41x      &{\bf [146]}-{\bf [70]}-[10]  &{\bf 492}&$\rightarrow$&{\bf 37} &4.88&1.95       \\
			SeReNe (local)                                           &1.7       &19.94     &63.59      &25.07x      &[192]-[83]-[10]              &656&$\rightarrow$&70       &3.83&1.93       \\
			\hline
	\end{tabular}
\end{table*}

%% file: tables/lenet5_mnist_tab.tex
\begin{table*}[!t]
	\renewcommand{\arraystretch}{1.3}
	\caption{LeNet-5 trained on MNIST.}
	\label{tab:lenet5}
	\centering
		\begin{tabular}{c c c c c c c c c c c}
			\hline
			\multirow{2}{*}{Approach} & \multicolumn{4}{c}{Remaining parameters (\%)} & Compr.& \multirow{2}{*}{Neurons} & \multicolumn{3}{c}{Network size [kB]}& Top-1\\
			& Conv1 & Conv2 & FC1 & FC2&ratio&& \texttt{.onnx} & & \texttt{.7z} &(\%) \\
			\hline
			Baseline                                                    &100          &100      &100          &100      &1x          &[20]-[50]-[500]-[10]                   &1686       &$\rightarrow$&1510       & \textbf{0.68} \\
			Sparse VD \cite{molchanov:var_dropout}                      &33         &\bf{2} &\bf{0.2}   &5      &\bf{280x}   &-                                      &&-&& 0.75 \\ 
			Han~\textit{et al.}~\cite{han:deep_compression}             &66         &12     &8          &19     &11.9x       &-                                      &&-&& 0.77\\ 
			SWS \cite{ullrich:weight_sharing}                           &-          &-      &-          &-      &162x       &-                                      &&-&& 0.97 \\ 
			Tartaglione~\textit{et al.}~\cite{tartaglione:sensitivity}  &67.6       &11.8   &0.9        &31.0   &51.1x       &[20]-[48]-[344]-[10]                   &&-&& 0.78 \\ 
			DNS~\cite{guo:surgery}                                      &\bf{14}    &3      &0.7        &\bf{4} &111x        &-                                      &&-&& 0.91 \\ 
			$\ell_2$+pruning                                               &60.20      &7.37   &0.61       &22.14  &72.3     &[19]-[37]-[214]-[10]    &577&$\rightarrow$&46&0.8 \\
			SeReNe (LB)                                               &33.75      &3.25   &0.27       &10.22  &177.05x     &\bf{[11]}-\bf{[26]}-\bf{[113]}-[10]    &{\bf 208}  &$\rightarrow$&{\bf 19}   &0.8 \\
			\hline
	\end{tabular}
\end{table*}

%% file: tables/lenet5_fmnist_tab.tex
\begin{table*}
	\renewcommand{\arraystretch}{1.3}
	\caption{LeNet-5 trained on Fashion-MNIST.}
	\label{tab:lenet5fashion}
	\centering
		\begin{tabular}{c c c c c c c c c c c}
			\hline
			\multirow{2}{*}{Approach} & \multicolumn{4}{c}{Remaining parameters (\%)} & Compr. & \multirow{2}{*}{Neurons} & \multicolumn{3}{c}{Network size [kB]}& Top-1\\
			& Conv1 & Conv2 & FC1 & FC2&ratio&& \texttt{.onnx} & & \texttt{.7z} &(\%) \\
			\hline
			Baseline                                                    &100          &100          &100          &100          &1x          &[20]-[50]-[500]-[10]       &1686       &$\rightarrow$&1510       &{\bf 8.1} \\
			Tartaglione~\textit{et al.}~\cite{tartaglione:sensitivity}  &{\bf 76.2}  &32.56      &6.5        &{\bf 44.02}&11.74x      &[20]-{\bf[47]}-[470]-[10]  &&-          &&          8.5 \\ 
			$\ell_2$+pruning                                               &85.80      &34.13      &4.57  &55.24      &14.36x      &[20]-[50]-[500]-[10]  &1496&$\rightarrow$&197   &8.44\\
			SeReNe (LB)                                               &85.71      &32.14      &{\bf 3.63}  &52.03      &17.04x      &[20]-[49]-{\bf [449]}-[10]  &{\bf 1494} &$\rightarrow$&{\bf 46}   &8.47\\
			\hline
	\end{tabular}
\end{table*}

%% file: tables/vgg1_cifar10_tab.tex
\begin{table*}
\renewcommand{\arraystretch}{1.3}
\caption{VGG-like architecture with 1 fully connected layer (\textit{VGG-1}) trained on CIFAR-10.}
\label{tab:VGG1}
\centering
   \begin{tabular}{c c c c c c c c c c c c}
    \hline
     \multirow{2}{*}{Approach} & \multicolumn{6}{c}{Remaining parameters (\%) [neurons]} & Compr. & \multicolumn{3}{c}{Network size [MB]}& Top-1\\
    & Conv1&Conv2&Conv3&Conv4&Conv5& FC1 &ratio& \texttt{.onnx} & & \texttt{.7z} &(\%)\\
  
    \hline
    Baseline&100&100&100&100&100&-&1x&57.57&$\rightarrow$&51.51&\textbf{7.36}\\
        &[64] & [128] & [256] & [512] & [512] & [10]\\
        &[64] & [128] & [256] & [512] & [512] \\
        &     &       & [256] & [512] & [512] \\
    $\ell_2$+pruning
        &11.86&15.07& 6.59&0.36&0.11&66.70
        &88.84x&13.58&$\rightarrow$&1.14&7.79\\
        
        &[23]&[126]&[250]&[406]&\textbf{[60]}&[10]\\
        &[64]&[123]&[251]&[108]&[81]\\
        &    &     &[250]&\textbf{[128]}&[398]\\
    SeReNe (LB)
        &10.18&11.68&\textbf{4.73}&\textbf{0.20}&\textbf{0.05}&\textbf{61.11}
        &\textbf{124.82x}&{\bf 11.56}&$\rightarrow$&{\bf 0.97}&7.8\\
        
        &[23]&[126]&[250]&\bf{[382]}&[65]&[10]\\
        &[64]&[123]&[251]&\bf{[93]}&\bf{[76]}\\
        &    &     &[250]&[136]&\bf{[373]}\\
        
        
    \hline
  \end{tabular}
\end{table*}

%% file: tables/vgg2_cifar10_tab.tex
\begin{table*}
\renewcommand{\arraystretch}{1.3}
\caption{VGG-like architecture with 2 fully connected layers (\textit{VGG-2}) trained on CIFAR-10.}
\label{tab:VGG2}
\centering
   \begin{tabular}{c c c c c c c c c c c c c}
    \hline
     \multirow{2}{*}{Approach} & \multicolumn{7}{c}{Remaining parameters (\%) [neurons]} & Compr. & \multicolumn{3}{c}{Network size [MB]}& Top-1\\
    & Conv1&Conv2&Conv3&Conv4&Conv5& FC1 & FC2 &ratio& \texttt{.onnx} & & \texttt{.7z} &(\%)\\
    \hline
    Baseline& 
          100 & 100 & 100 & 100 & 100 & 100 & 100 & 1x &58.61&$\rightarrow$&52.44&\textbf{6.16}\\
        &[64] & [128] & [256] & [512] & [512] & [512] & [10]\\
        &[64] & [128] & [256] & [512] & [512] \\
        &     &       & [256] & [512] & [512] \\
    Sparse-VD \cite{molchanov:var_dropout} & - & - & - & - & - & - & - & 48x &&-&& 7.3 \\
    
    $\ell_2$+pruning
        &27.62&30.74&13.67&0.88&0.24&1.88&70.78
        &40.96x&34.42&$\rightarrow$&2.86&7.21\\
        
        &[44] & [126] & [247] & [498] & [409] & [367] &[10] \\
        &[60] & [120] & [247] & [463] & [417] \\
        &     &       & [243] & [79]  & [461] \\
        
    SeReNe (LB)
        &\textbf{25.9}&\textbf{26.38}&\textbf{9.75}&\textbf{0.48}&\textbf{0.15}&\textbf{1.24}&\textbf{70}
        &\textbf{57.99x}&{\bf 29.41}&$\rightarrow$&{\bf 2.47}&7.25\\
        
        &[44] & [126] & [247] & [498]      & \bf{[354]} & [367] &[10] \\
        &[60] & [120] & [247] & \bf{[433]} & \bf{[366]} \\
        &     &       & [243] & \bf{[65]}  & \bf{[459]} \\
        
        
    \hline
  \end{tabular}
\end{table*}

%% file: tables/resnet32_cifar10.tex
\begin{table*}
\renewcommand{\arraystretch}{1.3}
\caption{ResNet-32 trained on CIFAR-10.}
\label{tab:resnet32}
\centering
    \begin{tabular}{c c c c c c c c c c c}
        \hline
        \multirow{2}{*}{Approach} & \multicolumn{5}{c}{Remaining parameters (\%) [neurons]} & Compr. & \multicolumn{3}{c}{Network size [MB]}& Top-1\\
         &Conv1&Block1&Block2&Block3&FC1&ratio & \texttt{.onnx} & & \texttt{.7z} &(\%)\\
        \hline
        Baseline &100 &100 &100 &100 &100 &1x &1.84 &$\rightarrow$ &1.63 &\textbf{7.36}\\
        &[64]   &[160]  &[320]  &[640]  &[10]\\
        $\ell_2$+pruning &65.97 &33.30 &33.41 &26.32 &88.75 &3.51x &1.82&$\rightarrow$ &0.54 &8.08\\
        &[14]   &[157]  &[319]  &[633]  &[10]\\
        SeReNe (LB) &\textbf{60.19} &\textbf{24.52} &\textbf{24.14} &\textbf{17.84} &\textbf{81.88} &\textbf{5.03x} &\textbf{0.87} &$\rightarrow$ &\textbf{0.37} &8.09\\
        &[\textbf{12}]   &[\textbf{93}]  &[\textbf{203}]  &[\textbf{364}]  &[10]\\
        \hline
  \end{tabular}
\end{table*}

%% file: tables/alexnet_cifar100_tab.tex
\begin{table*}
\renewcommand{\arraystretch}{1.3}
\caption{AlexNet trained on CIFAR-100.}
\label{tab:Alexnet}
\centering
    \begin{tabular}{c c c c c c c c c c c c c c c}
        \hline
        \multirow{2}{*}{Approach} & \multicolumn{8}{c}{Remaining parameters (\%) [neurons]} & Compr. & \multicolumn{3}{c}{Network size [MB]}& Top-1 & Top-5\\
         &Conv1&Conv2&Conv3&Conv4&Conv5&FC1&FC2&FC3&ratio& \texttt{.onnx} & & \texttt{.7z} &(\%)&(\%)\\
        \hline
        Baseline    &100 &100 &100 &100 &100 &100 &100 &100 &1x &92.31&$\rightarrow$&79.27&45.58 &20.09\\
        &[64]   &[192]  &[384]  &[256]  &[256]  &[4096] &[4096] &[100]\\
        $\ell_2$+pruning    &75.00 &21.95 &5.21 &3.65 &5.59 &0.62 &0.17 &6.44 &114.45x &60.88&$\rightarrow$&3.56&46.43&19.91\\
        &[64]   &[192]  &[384]  &[256]  &[256]  &[4094] &[2180] &[100]\\
        
        SeReNe (LB) & \textbf{79.05}& \textbf{20.33} &\textbf{5.72} &\textbf{3.33}&\textbf{2.23}&\textbf{0.18} &\textbf{0.04} &\textbf{2.77}&\textbf{179.52x}&{\bf 43.80}&$\rightarrow$&{\bf 2.47} &\textbf{44.99} &\textbf{17.88}\\
        &[64] &[\textbf{191}] &[384] &[256] &[256] &[\textbf{3322}] &[\textbf{1310}] &[100]\\
        \hline
  \end{tabular}
\end{table*}

%

%% file: tables/resnet101_imagenet_tab.tex
\begin{table*}
\renewcommand{\arraystretch}{1.3}
\caption{ResNet-101 trained on ImageNet.}
\label{tab:resnet101}
\centering
    \begin{tabular}{c c c c c c c c c c c c c}
        \hline
        \multirow{2}{*}{Approach} & \multicolumn{6}{c}{Remaining parameters (\%) [neurons]} & Compr. & \multicolumn{3}{c}{Network size [kB]}& Top-1 & Top-5\\
         &Conv1&Block1&Block2&Block3&Block4&FC1&ratio& \texttt{.onnx} & & \texttt{.7z} &(\%)&(\%)\\
        \hline
        Baseline  &100 &100 &100 &100 &100 &100 &1x &174.49 &$\rightarrow$ &156.67 &\textbf{22.63} &\textbf{6.44}\\
        &[64]   &[1408]  &[3584]  &[36352]  &[11264]  &[1000]\\
        $\ell_2$+pruning  &\textbf{53.12} &25.42&25.57&13.71&17.74&51.94&5.75x &172.94&$\rightarrow$ &32.93 &28.33&9.18\\
        &[49]   &[1241]  &[3280]  &[33278]  &[11250]  &[1000]\\
        SeReNe (LB) &55.36 &\textbf{24.27} &\textbf{23.79} &\textbf{11.24} &\textbf{14.81} &\textbf{40.82} &\textbf{6.94}x &\textbf{172.15} &$\rightarrow$ &\textbf{27.84} &28.41 &9.45\\
        &[49] &[\textbf{1197}] &[\textbf{3142}] &[\textbf{31948}] &[\textbf{11249}] &[1000]\\
        \hline
  \end{tabular}
\end{table*}

%% file: 6_conclusion.tex
\section{Conclusions}
\label{sec:conclusion}

In this work we have proposed a sensitivity-driven neural regularization technique. The effect of this regularizer is to penalize all the parameters belonging to a neuron whose output is not influential in the output of the network.
We have learned that the evaluation of the sensitivity at the neuron level (SeReNe) is extremely important in order to promote a structured sparsity in the network, being able to obtain a smaller network with minimal performance loss.
Our experiments show that the SeReNe strikes a favorable trade-off between ability to prune neurons and computational cost, while controlling the impairment in classification performance. 
For all the tested architectures and datasets, our sensitivity-based approach proved to introduce a structured sparsity while achieving state-of-the-art compression ratios. Furthermore, the designed sparsifying algorithm, making use of cross-validation, guarantees minimal (or no) performance loss, which can be tuned by the user via an hyper-parameter (TWT).
Future work includes deployment on physical embedded devices making use of deep network as well as using a quantization-based regularization jointly with the neuron sensitivity to further compress deep networks.

%% file: A1_upperboundeff.tex
\section{J made explicit}
\label{app:SUP}
In order to compute $S$ given (5), we can proceed directly. However, this is problematic as it requires $C$ different calls for the differentiation engine. Let us recall (4): 
\begin{equation*}
    S_{n,i} = \frac{1}{C}\sum_{k=1}^C \left| \frac{\partial y_{N, k}}{\partial p_{n,i}}\right|.
\end{equation*}
Now we inspect whether we can reduce the computation by defining an overall objective function which, when differentiated, yields $S$ as a result. We name it $J$:
\begin{align}
    J &=\int S_{n,i} dp_{n,i} = \int \frac{1}{C}\sum_{k=1}^C \left| \frac{\partial y_{N, k}}{\partial p_{n,i}}\right|dp_{n,i}=\nonumber\\
    &=\frac{1}{C}\int \sum_{k=1}^C \frac{\partial y_{N, k}}{\partial p_{n,i}} \sign\left(\frac{\partial y_{N, k}}{\partial p_{n,i}} \right) dp_{n,i}.
\end{align}
Here we can use Fubini-Tonelli's theorem:
\begin{align}
    J &=\frac{1}{C} \sum_{k=1}^C \int \frac{\partial y_{N, k}}{\partial p_{n,i}} \sign\left(\frac{\partial y_{N, k}}{\partial p_{n,i}} \right) dp_{n,i}\nonumber \\
    &=\frac{1}{C} \sum_{k=1}^C y_{N,k} \sign\left(\frac{\partial y_{N, k}}{\partial p_{n,i}} \right).
\end{align}
Unfortunately, we have no efficient way to compute $ \sign\left(\frac{\partial y_{N, k}}{\partial p_{n,i}} \right)$ and the only certain way is to compute $\frac{\partial y_{N, k}}{\partial p_{n,i}}$ directly, $\forall~C$.  

%% file: tables/lenet_300_sigmoid_tab.tex
\section{LeNet300 with sigmoid activation on MNIST}
\begin{table*}[tb]
	\renewcommand{\arraystretch}{1.3}
	\caption{LeNet-300 trained on MNIST (sigmoid activation, 1.7\% error rate).}
	\label{tab:lenet300_17_sig}
	\centering
		\begin{tabular}{c c c c c c c c c c}
			\hline
			\multirow{2}{*}{Approach} & \multicolumn{3}{c}{Remaining parameters (\%)} & Compr. & Remaining & \multicolumn{3}{c}{Model size [kB]} &  Top-1\\
			& FC1 & FC2 & FC3 & ratio & neurons & \texttt{.onnx} & & \texttt{.7z} & (\%)\\
			\hline
			Baseline                                                   &100         & 100        & 100         & 1x         &[300]-[100]-[10]             &1043&$\rightarrow$&963     &1.72 \\
			$\ell_2$ + pruning                                         &46.27     &82.87     &97.90      &1.97x      &[300]-[100]-[10]              &1043&$\rightarrow$&536   &1.75\\
			SeReNe                                                   &{\bf 2.44}&{\bf 16.73}&{\bf 85.30}&{\bf 22.31x}      &{\bf[215]}-[100]-[10]        &{\bf 749}&$\rightarrow$&{\bf 75}       &1.72       \\
			\hline
	\end{tabular}
\end{table*}
\begin{table*}[tb]
	\renewcommand{\arraystretch}{1.3}
	\caption{LeNet-300 trained on MNIST (sigmoid activation, 1.95\% error rate).}
	\label{tab:lenet300_19_sig}
	\centering
		\begin{tabular}{c c c c c c c c c c}
			\hline
			\multirow{2}{*}{Approach} & \multicolumn{3}{c}{Remaining parameters (\%)} & Compr. & Remaining & \multicolumn{3}{c}{Model size [kB]} & Top-1\\
			& FC1 & FC2 & FC3 & ratio & neurons & \texttt{.onnx} & & \texttt{.7z} & (\%)\\
			\hline
			Baseline                                                   &100         & 100        & 100         & 1x         &[300]-[100]-[10]             &1043&$\rightarrow$&963     &{\bf1.72} \\
			$\ell_2$ + pruning                                         &4.32      &30.53     &90.80      &12.92x      &[290]-[100]-[10]              &1008&$\rightarrow$&112    &1.98\\
			SeReNe                                                   &{\bf 1.15}&{\bf 9.32} &{\bf 76.00}      &{\bf 40.66x}      &{\bf[179]}-{\bf [99]}-[10]  &{\bf 624} &$\rightarrow$&{\bf 45}      &1.95       \\     \hline 
	\end{tabular}
\end{table*}

%% file: A2_Jderivation.tex
\section{Explicit derivation for SeReNe regularization function}
\label{app:Jderiv}
Here we focus on the update rule~\eqref{eq:updaterule}: we aim at minimizing the overall objective function
\begin{equation}
    O = \eta L + \lambda R,
\end{equation}
where
\begin{equation}
    R = \int w_{n,i,j} \bar{S}_{n,i} dw_{n,i,j}.
    \label{eq:R1}
\end{equation}
Here on we will drop the subscripts $n,i,j$. Let us consider the formulation of the sensitivity in \eqref{eq:3}: \eqref{eq:R1} becomes
\begin{equation}
    \label{eq:R2}
    R = \int w\cdot \bar{S}\cdot \Theta\left(\bar{S}\right)\cdot dw
\end{equation}
where $\Theta(\cdot)$ is the one-step function and
\begin{equation}
    \bar{S} = 1 - \frac{1}{C}\sum_{k=1}^{C} \left | \frac{\partial y_{N, k}}{\partial p}\right |.
\end{equation}
We can re-write \eqref{eq:R2} as
\begin{align}    
    R &= \int w \left[1 - \frac{1}{C}\sum_{k=1}^{C} \left | \frac{\partial y_{N, k}}{\partial p}\right | \right] \Theta\left(\bar{S}\right) dw \nonumber\\
    &= \int w\cdot\Theta\left(\bar{S}\right)\cdot dw - \int w\cdot \frac{1}{C}\sum_{k=1}^{C} \left | \frac{\partial y_{N, k}}{\partial p}\right |\cdot\Theta\left(\bar{S}\right)\cdot dw \nonumber\\
    &= \frac{w^2}{2}\cdot\Theta\left(\bar{S}\right) - \int w\cdot \frac{1}{C}\sum_{k=1}^{C} \left | \frac{\partial y_{N, k}}{\partial p}\right |\cdot\Theta\left(\bar{S}\right)\cdot dw.
\end{align}
Let us define
\begin{equation}
    \label{eq:J1}
    J_R= - \int w\cdot \frac{1}{C}\sum_{k=1}^{C} \left | \frac{\partial y_{N, k}}{\partial p}\right | \cdot\Theta\left(\bar{S}\right)\cdot dw.
\end{equation}
Considering that $w$ is $k$-independent and that $\frac{1}{C}$ is a constant, we can write
\begin{equation}
    \label{eq:J2}
    J_R= - \frac{1}{C} \int \sum_{k=1}^{C} w \left | \frac{\partial y_{N, k}}{\partial p}\right | \cdot\Theta\left(\bar{S}\right)\cdot dw.
\end{equation}
Here we are allowed to apply Fubini-Tonelli's theorem, swapping sum and integral:
\begin{align}
    \label{eq:J3}
    J_R&= - \frac{1}{C} \sum_{k=1}^{C} \int w \left | \frac{\partial y_{N, k}}{\partial p}\right | \cdot\Theta\left(\bar{S}\right)\cdot dw\nonumber\\
    &= - \frac{1}{C} \sum_{k=1}^{C} \int w  \frac{\partial y_{N, k}}{\partial p}\cdot \sign\left(\frac{\partial y_{N, k}}{\partial p} \right) \cdot\Theta\left(\bar{S}\right)\cdot dw.
\end{align}
Now integrating by parts:
\begin{align}
    J_R&=& - \frac{1}{C} \sum_{k=1}^{C} \int w  \frac{\partial y_{N, k}}{\partial p}\cdot \sign\left(\frac{\partial y_{N, k}}{\partial p} \right) \cdot\Theta\left(\bar{S}\right)\cdot dw\nonumber\\
    &=& - \frac{1}{C} \sum_{k=1}^{C} \left\{\frac{w^2}{2} \frac{\partial y_{N, k}}{\partial p}\cdot \sign\left(\frac{\partial y_{N, k}}{\partial p} \right) \cdot\Theta\left(\bar{S}\right)+\right .\nonumber \\
    &&\left .+ \int \frac{w^2}{2}\cdot \frac{\partial}{\partial w} \frac{\partial y_{N, k}}{\partial p}\cdot \sign\left(\frac{\partial y_{N, k}}{\partial p} \right) \cdot\Theta\left(\bar{S}\right)\cdot dw\right\}. \label{eq:parts1}
\end{align}
According to the derivative chain rule, we can re-write \eqref{eq:parts1} as
\begin{align}
   J_R =& - \frac{1}{C} \sum_{k=1}^{C} \left\{\frac{w^2}{2} \frac{\partial y_{N, k}}{\partial p}\cdot \sign\left(\frac{\partial y_{N, k}}{\partial p} \right) \cdot\Theta\left(\bar{S}\right)+\right .\nonumber \\
    &\left .+ \int \frac{w^2}{2}\cdot \frac{\partial^2 y_{N, k}}{\partial p^2}\cdot \frac{\partial p}{\partial w} \cdot \sign\left(\frac{\partial y_{N, k}}{\partial p} \right) \cdot\Theta\left(\bar{S}\right)\cdot dw\right\}. \label{eq:parts2}  
\end{align}
Applying infinite steps of integration by parts we have in the end
\begin{align}
    J_R =& \frac{1}{C} \Theta\left(\bar{S}\right) \sum_{k=1}^{C} \sign\left(\frac{\partial y_{N, k}}{\partial p}\right) \left[ -\frac{w^2}{2}\frac{\partial y_{N, k}}{\partial p} +\right . \nonumber\\
    &\left . +\sum_{i=1}^{\infty}(-1)^{i+1} \frac{w^{i+2}}{(i+2)!}\frac{\partial^{i+1} y_{N, k}}{\partial p^{i+1}}\frac{\partial^{i} p}{\partial w^{i}}\right].
\end{align}
Hence, the overall minimized $R$ function is
\begin{align}
    R = &\Theta\left(\bar{S}\right) \left\{\frac{w^2}{2} + \frac{1}{C} \sum_{k=1}^{C} \sign\left(\frac{\partial y_{N, k}}{\partial p}\right) \left[ -\frac{w^2}{2}\frac{\partial y_{N, k}}{\partial p} +\right . \right .\nonumber\\
    &\left .\left . +\sum_{i=1}^{\infty}(-1)^{i+1} \frac{w^{i+2}}{(i+2)!}\frac{\partial^{i+1} y_{N, k}}{\partial p^{i+1}}\frac{\partial^{i} p}{\partial w^{i}}\right] \right\}.
\end{align}

%% file: A3_17deriv.tex
\section{Derivation of \eqref{eq:updaterule2}}
\label{app:17deriv}
Let us recall the formulation in \eqref{eq:updaterule}. According to \eqref{eq:insensitivity}, we can re-write it as
\begin{equation}
    w_{n,i,j}^{t+1} = w_{n,i,j}^{t} - \eta \frac{\partial L}{w_{n,i,j}^{t}} - \lambda w_{n,i,j} (1-S_{n,i}) \Theta (1-S_{n,i}).
\end{equation}
Given the definition of $S_{n,i}$ in \eqref{eq:3}, we can write
\begin{align}
    w_{n,i,j}^{t+1} =& w_{n,i,j}^{t} - \eta \frac{\partial L}{w_{n,i,j}^{t}} +\nonumber\\
    &- \lambda w_{n,i,j} \Theta (1-S_{n,i}) \left(1-\frac{1}{C}\sum_{k=1}^{C} \left | \frac{\partial y_{N, k}}{\partial p_{n,i}}\right |\right).
\end{align}
We can multiply the insensitivity by the term $w_{n,i,j}$:
\begin{align}
    w_{n,i,j}^{t+1} =& w_{n,i,j}^{t} - \eta \frac{\partial L}{w_{n,i,j}^{t}} +\nonumber\\
    &- \lambda \Theta (1-S_{n,i}) \left(w_{n,i,j}-\frac{1}{C}\sum_{k=1}^{C} \left | \frac{\partial y_{N, k}}{\partial p_{n,i}}\right |\cdot w_{n,i,j}\right).
\end{align}
Finally here, observing \eqref{eq:derivisw}, we find back \eqref{eq:updaterule2}.

%% file: A4_sigmoidact.tex
\section{LeNet300 with sigmoid activation on MNIST}
\label{sec:ln300sigmoid}

Finally, we repeat the experiment in Sec.~\ref{sec:ln300} yet replacing the ReLU activations with sigmoids in the hidden layers. 
We optimize a pre-trained LeNet300 using SGD with learning rate $\eta = 0.1$, $PWE=20$ epochs, $TWT = 0$ for target Top-1 errors of 1.7\% (Table~\ref{tab:lenet300_17_sig}) and 1.95\% (Table~\ref{tab:lenet300_19_sig}).\\
SeReNe achieves a sparser (and smaller) architecture than $\ell_2$~+~pruning for both error rates. Interestingly, for the 1.7\% error rate, $\ell_2$~+~pruning is not able to prune any neuron, whereas SeReNe prunes 85 neurons from FC1, with a 10 times higher compression ratio. This reflects in the compressed model size: while $\ell_2$~+~pruning squeezes the architecture to 536kB, SeReNe compresses it to 75kB only.